\documentclass[11pt,a4paper]{article}
\usepackage[hyperref]{acl2020}
\usepackage{times}
\usepackage{placeins}
\usepackage{graphicx}
\usepackage{booktabs}
\usepackage{latexsym}

\usepackage{microtype}

\aclfinalcopy 
\def\aclpaperid{55} 

\title{Cognitively Aided Zero-Shot Automatic Essay Grading}

\author{Sandeep Mathias$^{1}$, Rudra Murthy$^{1,2}$, Diptesh Kanojia$^{1,3}$, \and Pushpak Bhattacharyya$^{1}$ \\
$^{1}$ Department of Computer Science \& Engineering, IIT Bombay \\
$^{2}$ IBM Research India Limited \\
$^{3}$ IITB-Monash Research Academy \\
\texttt{\{sam,diptesh,pb\}@cse.iitb.ac.in, rmurthyv@in.ibm.com} \\}

\date{}

\begin{document}
\maketitle
\begin{abstract}
Automatic essay grading (AEG) is a process in which machines assign a grade to an essay written in response to a topic, called the prompt. Zero-shot AEG is when we train a system to grade essays written to a new prompt which was not present in our training data. In this paper, we describe a solution to the problem of zero-shot automatic essay grading, using cognitive information, in the form of gaze behaviour. Our experiments show that using gaze behaviour helps in improving the performance of AEG systems, especially when we provide a new essay written in response to a new prompt for scoring, by an average of almost \textbf{5 percentage points} of QWK.
\end{abstract}

\section{Introduction}
\label{Introduction Section}

One of the major challenges in machine learning is the requirement of a large amount of training data. AEG systems perform at their best when they are trained in a prompt-specific manner - i.e. the essays that they are tested on are written in response to the \textbf{same} prompt as the essays they are trained on \cite{zesch-etal-2015-task}. These systems perform badly when they are tested against essays written in response to a different prompt.

Zero-shot AEG is when our AEG system is used to grade essays written in response to a completely different prompt. In order to solve this challenge of lack of training data, we use cognitive information learnt by gaze behaviour of readers to augment our training data and improve our model.

Automatic essay grading has been around for over half a century ever since \newcite{page1966imminence}'s work \cite{beigman-klebanov-madnani-2020-automated}. While there have been a number of commercial systems like E-Rater \cite{Attali06jtla} from the Educational Testing Service (ETS), most modern-day systems use deep learning and neural networks, like convolutional neural networks \cite{dong-zhang-2016-automatic}, recurrent neural networks \cite{taghipour-ng-2016-neural}, or both \cite{dong-zhang-2016-automatic}. However, all these systems rely on the fact that their training and testing data is from the same prompt.

Quite often, at run time, we may not have essays written in response to our target prompt (i.e. the prompt which our essay is written in response to). Because of the lack of training data, especially when training a model for essays written for a new prompt, many systems may fail at run time. To solve this problem, we propose a multi-task approach, similar to \newcite{mathias2020happy}, where we learn a reader's gaze behaviour for helping our system grade new essays.

In this paper, we look at a similar approach proposed by \newcite{mathias2020happy} to grade essays using cognitive information, which is learnt as an auxiliary task in a multi-task learning approach. Multi-task learning is a machine-learning approach, where the model tries to solve one or more auxiliary tasks to solve a primary task \cite{Caruana1998}. Similar to \newcite{mathias2020happy}, the scoring of the essay is the primary task, while learning the gaze behaviour is the auxiliary task.

\paragraph{Contribution.} In this paper, we describe a relatively new problem - zero-shot automatic essay grading - and propose a solution for it using gaze behaviour data. We show a \textbf{5 percentage points} increase in performance when learning gaze behaviour, as opposed to without using it.

\subsection{Gaze Behaviour Terminology}
\label{Terminology Subsection}

We use the following gaze behaviour terms as defined by \newcite{mathias2020happy}. An \textbf{\textit{Interest Area}} (IA) is a part of the screen that is of interest to us. These areas are where some text is displayed, and not the background on the left / right, as well as above / below the text. \textbf{Each word} is a separate and unique IA. A \textbf{\textit{Fixation}} is an event when the reader's eye fixates on a part of the screen. We are only concerned with fixations that occur inside interest areas. The fixations that occur in the background are ignored. \textbf{\textit{Saccades}} are eye movements as the eye moves from one fixation point to the next. \textbf{\textit{Regressions}} are a type of saccade where the reader moves from the current interest area to an \textit{earlier} one.

\subsection{Organization of the Paper}
The rest of the paper is organized as follows. Section \ref{Motivation Section} describes the motivation for our work. Section \ref{Related Work Section} describes some of the related work in the area of automatic essay grading. Section \ref{Dataset Section} describes the essay dataset, as well as the gaze behaviour dataset. Section \ref{Experiment Section} describes our experiment setup. We report our results and analyze them in Section \ref{Results Section} and conclude our paper in Section \ref{Conclusion Section}.

\section{Motivation}
\label{Motivation Section}

As stated earlier, in Section \ref{Introduction Section}, one of the challenges for machine-learning systems is the requirement of training data. Quite often, we may not have training data for an essay, especially if the essay is written in response to a new prompt. Without any labeled data, in the form of scored essays, we cannot train a system properly to grade the essays.

Zero-shot automatic essay grading is a way in which we overcome this problem. In zero-shot automatic essay grading, we train our system on essays written to different prompts, and test it on essays written in response to the target prompt. One drawback of this approach is that it would not be able to use the properties of the target essay set in training the model. Therefore, as a way to alleviate this problem, we learn cognitive information, in the form of gaze behaviour, for the essays to help our automatic essay grading system grade the essays better.

\section{Related Work}
\label{Related Work Section}

While there has been work done on developing systems for automatic essay grading, all of them describe systems which use some of the essays the system is tested on as part of the training data (as well as validation data, where applicable) \cite{chen-he-2013-automated,phandi-etal-2015-flexible,taghipour-ng-2016-neural,dong-zhang-2016-automatic,dong-etal-2017-attention,zhang-litman-2018-co,cozma-etal-2018-automated,tay-2018-skipflow,mathias2020happy}.

One of the solutions to solve the problem was using cross-domain AEG, where systems were trained using essays in a set of source prompt / prompts and tested on essays written in response to the target prompt. Some of the work done to study cross-domain AEG were \newcite{zesch-etal-2015-task} (who used task-independent features),  \newcite{phandi-etal-2015-flexible} (who used domain adaptation), \newcite{dong-zhang-2016-automatic} (who used a hierarchical CNN layers) and \newcite{cozma-etal-2018-automated} (who used string kernels and super word embeddings). In all of their works, they defined a \textbf{\textit{source prompt}} which is used for training and a \textbf{\textit{target prompt}} which is used for validation and testing.

To the best of our knowledge, we are the first to explore the task of \textbf{\textit{Zero-shot}} automatic essay grading, as a way to alleviate the challenge of a lack of graded essays (written in response to the target prompt) for an automatic essay grading system. In our approach, \textbf{we do not use the target prompt essays even for validation}, thereby making it truly zero-shot.

\section{Datasets}
\label{Dataset Section}

In this section, we discuss our essay grading dataset and the gaze behaviour dataset which we used.

\subsection{Essay Dataset Details}

\begin{table*}[t]
\centering
\begin{tabular}{ccccc}
\toprule
\textbf{Prompt ID} & \textbf{Number of Essays} & \textbf{Score Range} & \textbf{Mean Word Count} & \textbf{Essay Type} \\ \midrule
Prompt 1 & 1783 & 2-12 & 350 & Persuasive \\ 
Prompt 2 & 1800 & 1-6 & 350 & Persuasive \\ \midrule
Prompt 3 & 1726 & 0-3 & 150 & Source-Dependent \\ 
Prompt 4 & 1770 & 0-3 & 150 & Source-Dependent \\ 
Prompt 5 & 1805 & 0-4 & 150 & Source-Dependent \\ 
Prompt 6 & 1800 & 0-4 & 150 & Source-Dependent \\ \midrule
Prompt 7 & 1569 & 0-30 & 250 & Narrative \\
Prompt 8 & 723 & 0-60 & 650 & Narrative \\ \midrule
\textbf{Total} & 12976 & 0-60 & 250 & -- \\ \bottomrule
\end{tabular}%
\caption{Statistics of the 8 prompts from the ASAP AEG dataset.}
\label{CD AEG Essay Dataset Details Table}
\end{table*}

For our experiments, we use the Automatic Student Assessment Prize (ASAP)'s AEG dataset\footnote{The dataset can by downloaded from \url{https://www.kaggle.com/c/asap-aes/data}.}. This dataset is one of the most widely-used essay grading datasets, consisting of 12,978 graded essays, written in response to 8 essay prompts. The prompts are either argumentative, narrative, and source dependent responses. Details of the dataset are summarized in Table \ref{CD AEG Essay Dataset Details Table}.

\subsection{Gaze Behaviour Dataset}

\begin{table}[h]
\centering
\begin{tabular}{ccccccc}
\toprule
\textbf{Essay Set} & \textbf{0} & \textbf{1} & \textbf{2} & \textbf{3} & \textbf{4} & \textbf{Total} \\ \midrule
Prompt 3 & 2 & 4 & 5 & 1 & N/A & 12 \\ 
Prompt 4 & 2 & 3 & 4 & 3 & N/A & 12 \\ 
Prompt 5 & 2 & 1 & 3 & 5 & 1 & 12 \\ 
Prompt 6 & 2 & 2 & 3 & 4 & 1 & 12 \\ \midrule
\textbf{Total} & 8 & 10 & 15 & 13 & 2 & 48 \\ \bottomrule
\end{tabular}%
\caption{Number of essays for each essay set which we collected gaze behaviour, scored between 0 to 3 (or 4).}
\label{Gaze Dataset Details Table}
\end{table}

For our experiments, we use the same essay grading dataset as \newcite{mathias2020happy}. We use 5 attributes of gaze behaviour, namely dwell time (the total time that the eye has fixated on a word), first fixation duration (the duration of the first fixation of the reader on a particular word), IsRegression (whether or not there was a regression from a particular interest area or not), Run Count (the number of times an interest area was fixated on), and Skip (whether or not the interest area was skipped).

\begin{table*}[t]
\centering
\resizebox{\textwidth}{!}{
\begin{tabular}{llclcccccc}
\toprule
\textbf{ID} & \textbf{Sex} & \textbf{Age} &  \textbf{Occupation} & \textbf{TA?} & \textbf{L1 Language} & \textbf{English Score} & \textbf{QWK} & \textbf{Correct} & \textbf{Close} \\
\hline
Annotator 1 & Male & 23 & Masters student & Yes & Hindi & 94\% & 0.611 & 19 & 41 \\
Annotator 2 & Male & 18 & Undergraduate & Yes & Marathi & 95\% & 0.587 & 24 & 41 \\
Annotator 3 & Male & 31 & Research scholar & Yes & Marathi & 85\% & 0.659 & 21 & 43 \\
Annotator 4 & Male & 28 & Software engineer & Yes & English & 96\% & 0.659 & 26 & 44 \\
Annotator 5 & Male & 30 & Research scholar & Yes & Gujarati & 92\% & 0.600 & 19 & 42 \\
Annotator 6 & Female & 22 & Masters student & Yes & Marathi & 95\% & 0.548 & 19 & 40 \\
Annotator 7 & Male & 19 & Undergraduate & Yes & Marathi & 93\% & 0.732 & 21 & 46 \\
Annotator 8 & Male & 28 & Masters student & Yes & Gujarati & 94\% & 0.768 & 29 & 45 \\
\hline
\end{tabular}
}
\caption{Profile of the annotators}
\label{Annotator Table}
\end{table*}

The gaze behaviour was collected from 8 different annotators, who read only 48 essays (out of the almost 13,000 essays in the ASAP AEG dataset) from the source dependent response essay sets.  Table \ref{Gaze Dataset Details Table} summarizes the distribution of essays across the different essay sets that we collect gaze behaviour data for.

Table \ref{Annotator Table} gives the details of the different annotators used by \newcite{mathias2020happy}. We evaluated the annotator's performance on 3 different metrics - QWK, Close and Correct. \textbf{QWK} is the Quadratic Weighted Kappa agreement \cite{cohen1968weighted} between the score given by the annotator and the ground truth score from the dataset. \textbf{Correct} is the number of times (out of 48) that the annotator \textbf{exactly} agreed with the ground truth score, and \textbf{Close} is the number of times (out of 48) where the annotator disagreed with the ground truth score by \textbf{at most 1 score point}.

More details about the dataset and its creation are found in \newcite{mathias2020happy}.


\section{Experiment Setup}
\label{Experiment Section}

In this section, we describe our experiment setup, such as the evaluation metric, network architecture and hyperparameters, etc.

\subsection{Evaluation Metric}

For evaluating our system, we use Cohen's Kappa with Quadratic Weights, i.e. Quadratic Weighted Kappa (QWK) \cite{cohen1968weighted}. This evaluation metric is most frequently used for automatic essay grading experiments because it is sensitive to differences in scores, and takes into account chance agreements \cite{mathias-etal-2018-eyes}.

\subsection{Network Architecture}

\begin{figure}[h]
\centering
\includegraphics[width=\columnwidth]{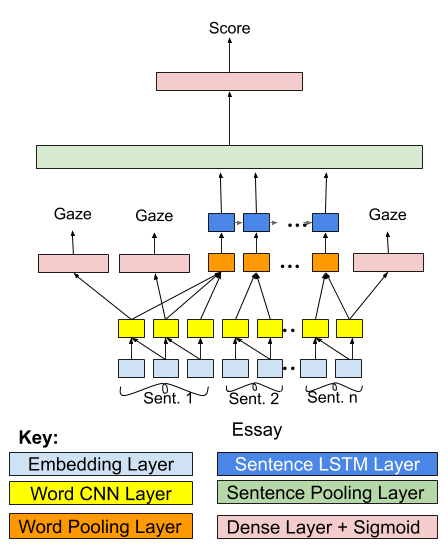}
\caption{Architecture of our gaze behaviour system, showing an input essay of $n$ sentences, with the outputs being the gaze behaviour (whenever applicable), and the overall essay score.}
\label{Network Architecture Figure}
\end{figure}

Figure \ref{Network Architecture Figure} shows the architecture of our system. The essay is split into different sentences and each sentence is tokenized and given as input at the Embedding Layer. In this layer, for each token, we output the corresponding word embedding, which is given as input to the next layer - the Word-level CNN layer.

The Word-level CNN layer learns local representations of nearby words, as well as the gaze behaviour. The outputs of the word-level CNN layer are then pooled at the word-level pooling layer to get a sentence representation for each sentence.

Each sentence representation is then sent through an LSTM \cite{hochreiter1997long} layer, whose output is pooled through a sentence-level attention layer, to get the essay representation.

The essay representation from the sentence-level attention layer is then sent through a Dense layer, from which we learn the essay scores. For both the tasks (learning gaze behaviour, as well as scoring the essay), we minimize the mean squared error loss.

\subsection{Network Hyperparameters}

We use the \textbf{50 dimension} GloVe pre-trained word embeddings \cite{pennington-etal-2014-glove}. We run our experiments over a \textbf{batch size of 200}, for \textbf{50 epochs}. We set the \textbf{learning rate as 0.001}, and the \textbf{dropout rate as 0.5}. The word-level CNN layer has a \textbf{kernel size of 5}, with \textbf{100 filters}. The sentence-level LSTM layer has \textbf{100 hidden units}. We use the RMSProp Optimizer \cite{dauphin2015rmsprop} with an \textbf{initial learning rate of 0.001} and \textbf{momentum of 0.9}. Along with the network hyperparameters, we also weigh the loss functions of the different gaze behaviour attributes differently, using the same weights as \newcite{mathias2020happy}, namely \textbf{0.05 for DT and FFD}, \textbf{0.01 for IR and RC}, and \textbf{0.1 for Skip}.

\subsection{Normalization and Binning}

While training our model, we scale the essay scores for all the data (training, testing and validation) to a range of $[0, 1]$. For calculating the final scores, as well as the QWK, we rescale the predictions of the essay score back to the score range of the essays.

We also bin the gaze behaviour attributes as described in \newcite{mathias2020happy}. Binning is done to take into account the idiosyncracies of the gaze behaviour of individual readers (i.e. some people may read faster, others slower, etc.). Whenever we use gaze behaviour, we scale the value of the gaze behaviour bins to the range of $[0,1]$ as well.

\subsection{Experiment Configurations}

We run our experiments in the following configurations. \textbf{No Gaze} is a single-task learning experiment, where we only learn to score the essay. \textbf{Gaze} is the multi-task learning approach, where we learn gaze behaviour as an auxiliary task, and score the essay as the primary task.

\subsection{Evaluation Method}

We use \textbf{five-fold cross-validation} to evaluate our system. For each fold, the testing data consists of essays from the target prompt and the training data and validation data comprise of essays from the other 7 prompts. 


\section{Results and Analysis}
\label{Results Section}

\begin{table}[h]
\centering
\begin{tabular}{|l|c|c|}
\hline
\textbf{Target Essay Set} & \textbf{No Gaze} & \textbf{Gaze} \\
\hline
Prompt 1 & 0.319 & \textbf{0.423*} \\
Prompt 2 & 0.391 & \textbf{0.439*} \\
Prompt 3 & 0.508 & \textbf{0.545*} \\
Prompt 4 & 0.548 & \textbf{0.626*} \\
Prompt 5 & 0.548 & \textbf{0.628*} \\
Prompt 6 & 0.599 & \textbf{0.600} \\
Prompt 7 & 0.362 & \textbf{0.420*} \\
Prompt 8 & \textbf{0.316} & 0.286 \\
\hline
\textbf{Mean QWK} & 0.449 & \textbf{0.498*} \\
\hline
\end{tabular}
\caption{Results of our experiments with and without using gaze behaviour. Improvements which are statistically significant (with $p < 0.05$), when gaze behaviour is used, are marked with a \textbf{*}}
\label{Results Table}
\end{table}

Table \ref{Results Table} gives the results of our experiments. The results reported are on the target essay set for the mean of the 5 folds. For each fold, we record the performance of the model on the target essay set, corresponding to the epoch which had the best QWK for the development set. Table \ref{Results Table} reports the mean performance for all 5 folds.

From the table, we see that in most of the essay sets, we are able to see an improvement in performance. In order to verify if the improvements were statistically significant, we use the 2-tailed Paired T-Test with a significance level of $p < 0.05$. Statistically significant improvements where we use gaze behaviour data are marked with a \textbf{*} next to the result.

Out of the 8 essay sets, the only essay set where the performance using gaze behaviour falls short compared to when we do not use gaze behaviour is in Prompt 8. One of the main reasons for this is that the essays in Prompt 8 are very long compared to the other essay sets. When they are absent from the training data, the system is unable to learn about the existence of long essays, which could also be the reason that those essays are scored badly.


\section{Conclusion and Future Work}
\label{Conclusion Section}

In this paper, we discussed an important problem for automatic essay grading, namely \textbf{zero-shot} automatic essay grading, where we have no labeled essays written in response to our target prompt, present at the time of training.

We showed that, by using gaze behaviour, we are able to learn cognitive information which can help improve our AEG system.

In the future, we plan to extend our work to other tasks, like grading of essay traits, using gaze behaviour.


\bibliography{anthology,acl2020}
\bibliographystyle{acl_natbib}

\end{document}